\documentclass{article}
\usepackage{float}
\usepackage{graphicx}
\usepackage{booktabs}
\usepackage{enumitem}
\usepackage{hyperref}
\usepackage{xcolor}
\usepackage{natbib}
\usepackage{tikz}
\usetikzlibrary{positioning,arrows.meta,calc}

\usepackage[final]{neurips_2026}
\makeatletter
\renewcommand{\@noticestring}{Workshop on Who Verifies the Agents? Toward Reliable Agent Development, NeurIPS 2026.}
\makeatother
\usepackage{amsmath}
\usepackage{amssymb}
\usepackage[capitalize,noabbrev]{cleveref}

\newcommand{\env}{\textsc{RetryRealistic}}

\title{When Tools Fail, Agents Fabricate:\\
  Silent Failure and Verification-First Oversight in Tool-Using LLM Agents}

\author{
  G\"un Kaynar \qquad Boshen Yan \qquad Carl Kingsford \\
  Carnegie Mellon University \\
  \texttt{carlk@cs.cmu.edu}
}

\begin{document}

\maketitle

\begin{abstract}
Tool-using LLM agents must recover when a tool call fails, by retrying,
switching tools, or revising their plan. We study how agents fail when
tools fail, using \env{}, a sandboxed yet realistic testbed for controlled
fault injection in which agents operate on a real OS filesystem with genuine permission
enforcement, SQLite databases, a deterministic local HTTP server, and a
constrained shell, across 8 tools, 9 fault types with controlled persistence,
and 30 multi-step tasks. Evaluating four agents over 2{,}100 faulted runs against ground truth, we
find that recovery is unreliable (19--68\% correct) and frequently
silent: 6--26\% of runs end in a confident fabrication, a wrong answer
asserted as fact. The failure mode is governed more by the fault type than by the
model: faults that return partial data drive fabrication (up to
38\%), while hard denials drive honest failure. The mechanism is what we call
\emph{silent ingestion}: the agent accepts faulted data as valid on first
contact without retrying or flagging the problem, then builds a
fabricated answer on top of it. We show that fabrication is reducible: a verification-first
Monitor, grounding checks in structured tool outputs and issuing corrective
directives rather than hard blocks, raises correct recovery from 45.6\% to
56.3\% and cuts fabrication from 17.3\% to 10.8\%. Removing chain-of-thought
increases fabrication ($+8.3$pp), so deliberation guards against silent
ingestion. We release \env{}, the fault-injection toolkit, and all annotated
traces.\footnote{\url{https://github.com/Kingsford-Group/RetryRealistic}}
\end{abstract}

\section{Introduction}
\label{sec:intro}

Tool-using LLM agents~\citep{schick2023toolformer,yao2022react} work by
calling external tools such as file systems, APIs, databases, and shells. In
practice, these calls can fail. Files lack permissions, databases return malformed
rows, HTTP services return 503s, and scripts exit with non-zero codes. An agent
that cannot recover from these failures cannot be trusted in production.

\textbf{Recovery is itself a source of failure.} When a tool call fails, an
agent must diagnose the error and adapt. But recovery is error-prone: agents
loop on persistent faults, drift from the original goal, or, at worst,
fabricate an answer, asserting a specific result they
never obtained. For a deployed agent, how it fails matters as much as
whether it fails: a loud, honest failure (``the data is unavailable'')
is recoverable downstream; a silent fabrication is not.

\textbf{Studying recovery needs a realistic substrate.} Whether an agent
fabricates or fails honestly depends on the specific content of the fault it
encounters (a real permission-denied kernel error, a truncated query result,
a locked database) and measuring fabrication requires grading answers against
ground truth. Simulated substrates (files as dictionaries, canned error
strings, no-op writes) produce neither the faults nor the ground truth needed.
We build a realistic yet controlled testbed and ask three questions:
(1)~When tools fail, do agents fail loudly or silently?
(2)~What determines the failure mode: the model, or the fault?
(3)~Can a second LLM, monitoring the agent, suppress silent failure?

We make the following contributions.
\begin{enumerate}[leftmargin=1.5em,itemsep=2pt]
  \item We introduce \textbf{\env{}}, a sandboxed, realistic fault-injection
  testbed that provides 8 tools over a real OS filesystem, SQLite, a local HTTP
  server, and a constrained shell, along with 9 fault types with controlled
  persistence and 30 multi-step tasks with designed alternative recovery paths
  and ground truth.

  \item \textbf{Silent failure is widespread.} Recovery is unreliable
  (19--68\% correct), and 6--26\% of faulted runs end in confident
  fabrication.

  \item \textbf{The fault type, more than the model, determines the failure mode.}
  Faults that return partial data drive fabrication (up to 38\%); hard denials
  drive honest failure. The mechanism is \emph{silent ingestion}: fabrication
  happens when the agent accepts faulted data on first contact without
  detecting the problem, before any retry or drift.

  \item \textbf{Fabrication is reducible.} A verification-first Monitor that
  grounds checks in structured tool outputs and issues corrective directives
  raises correct recovery and cuts fabrication. Chain-of-thought ablation and an honesty
  prompt probe further isolate the mechanism.

  \item \textbf{We identify a methodological issue.} LLM-judged behavioral taxonomies
  are unreliable (two judges agree at $\kappa\!=\!0.58$); an objective
  answer-vs-ground-truth metric is stable across judges ($\kappa\!=\!0.91$ on
  the same runs) and should be preferred.
\end{enumerate}

\section{Related Work}
\label{sec:related}

\textbf{Agent benchmarks} such as
WebArena~\citep{zhou2024webarena}, OSWorld~\citep{xie2024osworld},
AppWorld~\citep{trivedi2024appworld}, SWE-bench~\citep{jimenez2023swe},
WorkArena~\citep{drouin2024workarena}, AgentBench~\citep{liu2024agentbench},
GAIA~\citep{mialon2024gaia}, and $\tau$-bench~\citep{yao2024tau}
evaluate agents across web, OS, and tool-use settings, but none of them
controls fault type, persistence, or alternative recovery paths to isolate
recovery behavior as the dependent variable.

Several \textbf{fault injection} methods target agent systems.
MAS-FIRE~\citep{jia2026mas} injects cognitive and coordination faults into
multi-agent systems via prompt modification, response rewriting and message
routing; it does not study tool failures on a real substrate or control fault
persistence. ReliabilityBench~\citep{gupta2026reliabilitybench} evaluates LLM
agents under perturbations and injected tool faults and finds that a
self-reflecting architecture (Reflexion) is less robust than plain ReAct under
combined stress, consistent with our finding that reflection alone does not
make recovery reliable; it does not grade fabrication against ground truth or
propose a fix.
agent-chaos~\citep{BibEntry2026Aug} provides composable fault
injectors for practitioners but does not characterize failure patterns.
AgenTracer~\citep{zhang2026agentracer} uses fault injection for root-cause analysis of
individual trajectories.

\textbf{Error recovery} approaches such as
Reflexion~\citep{shinn2023reflexion} and Self-Refine~\citep{madaan2023self}
assume that LLMs improve by reflecting on prior outputs.
\citet{huang2024large} show that without external grounding,
self-correction often makes outputs worse. In our study, when a tool fails
loudly its error message is in the agent's context and agents still sometimes
fabricate; when it fails silently, as with a truncated result, there is no
error to reflect on at all.
SWE-Search~\citep{antoniades2025swe} applies Monte Carlo tree search to
software engineering tasks for structured backtracking. This motivates our
verification-first Monitor, which grounds recovery decisions in the agent's
tool outputs rather than in free-form reflection.

OpenHands~\citep{wang2025openhands} and BrowserGym~\citep{chezelles2024browsergym}
provide \textbf{sandboxed execution} environments for agent evaluation.
\env{} differs in that it is a fault-injection testbed, not a general scaffold,
with real OS calls for ecological validity and middleware-based injection for
control.

\section{The \env{} Environment}
\label{sec:env}

\begin{figure}[t]
\centering
\begin{tikzpicture}[
  font=\small,
  node distance=7mm and 10mm,
  box/.style={draw, rounded corners=3pt, minimum height=10mm, minimum width=22mm,
              align=center, inner sep=5pt, fill=gray!6, line width=0.4pt},
  sandboxbox/.style={draw, rounded corners=3pt, minimum height=14mm,
              minimum width=52mm, align=center, inner sep=5pt, fill=green!4,
              line width=0.6pt},
  agentbox/.style={box, fill=blue!6, minimum width=34mm, minimum height=13mm,
              line width=0.6pt},
  monbox/.style={box, fill=orange!8, minimum width=30mm, minimum height=22mm,
              line width=0.6pt},
  faultbox/.style={box, fill=red!6, line width=0.5pt},
  ar/.style={-{Latex[length=2mm, width=1.5mm]}, line width=0.65pt, color=gray!60},
  ardash/.style={-{Latex[length=2mm, width=1.5mm]}, line width=0.65pt, dashed,
              color=orange!50!black},
  lab/.style={font=\scriptsize\sffamily, fill=white, inner sep=1.5pt,
              text=gray!55!black},
]
  \node[box, minimum width=18mm] (yaml) {Task YAML};
  \node[box, right=10mm of yaml] (sbm) {Sandbox\\[-1pt]Manager};
  \node[sandboxbox, below=9mm of $(yaml.south east)!0.5!(sbm.south west)$]
    (sandbox) {\textbf{Sandbox}\, (fresh per run)\\[1pt]
    {\footnotesize real FS\;\,$\cdot$\;\,SQLite\;\,$\cdot$\;\,mock HTTP\;\,$\cdot$\;\,shell}};
  \node[box, below=9mm of sandbox] (reg) {Tool\\[-1pt]Registry};
  \node[faultbox, below=7mm of reg] (inj) {Fault\\[-1pt]Injector};
  \node[agentbox, below=10mm of inj] (agent)
    {ReAct Agent\\[1pt]
     {\footnotesize\textsf{THOUGHT\;$\to$\;ACTION\;$\to$\;OBS}}};
  \node[monbox, right=14mm of agent] (mon)
    {\textbf{Monitor}\\[2pt]
     {\footnotesize\textsf{S1 Grounding\, \textcolor{gray!70}{(code)}}}\\[-1pt]
     {\footnotesize\textsf{S2 Surface\, \textcolor{gray!70}{(code)}}}\\[-1pt]
     {\footnotesize\textsf{S3 Recompute\, \textcolor{gray!70}{(code)}}}\\[-1pt]
     {\footnotesize\textsf{S4 Drift\, \textcolor{gray!70}{(LLM)}}}};
  \draw[ar] (yaml) -- node[lab, above] {spec} (sbm);
  \draw[ar] (sbm) -- node[lab, right, pos=0.45] {materialize} (sandbox);
  \draw[ar] (reg) -- node[lab, right, pos=0.45] {execute} (sandbox);
  \draw[ar] (inj) -- (reg);
  \draw[ar] ([xshift=5mm]agent.north) --
    node[lab, right] {\textsf{ACTION}} ([xshift=5mm]inj.south);
  \draw[ar] ([xshift=-5mm]inj.south) --
    node[lab, left] {\textsf{OBS\;\textcolor{red!40!black}{/ fault}}}
    ([xshift=-5mm]agent.north);
  \draw[ardash] ([yshift=2mm]agent.east) --
    node[lab, above] {trace} ([yshift=2mm]mon.west);
  \draw[ardash] ([yshift=-2mm]mon.west) --
    node[lab, below] {\textsf{directive}} ([yshift=-2mm]agent.east);
\end{tikzpicture}
\caption{
  \env{} architecture. The Sandbox Manager materializes a fresh directory
  from the task YAML (files with real \texttt{chmod} permissions, SQLite
  databases, mock server routes). The Fault Injector intercepts the first call
  matching the task's \texttt{fault\_target}; subsequent calls follow the
  fault's persistence semantics.
}
\label{fig:env_overview}
\end{figure}
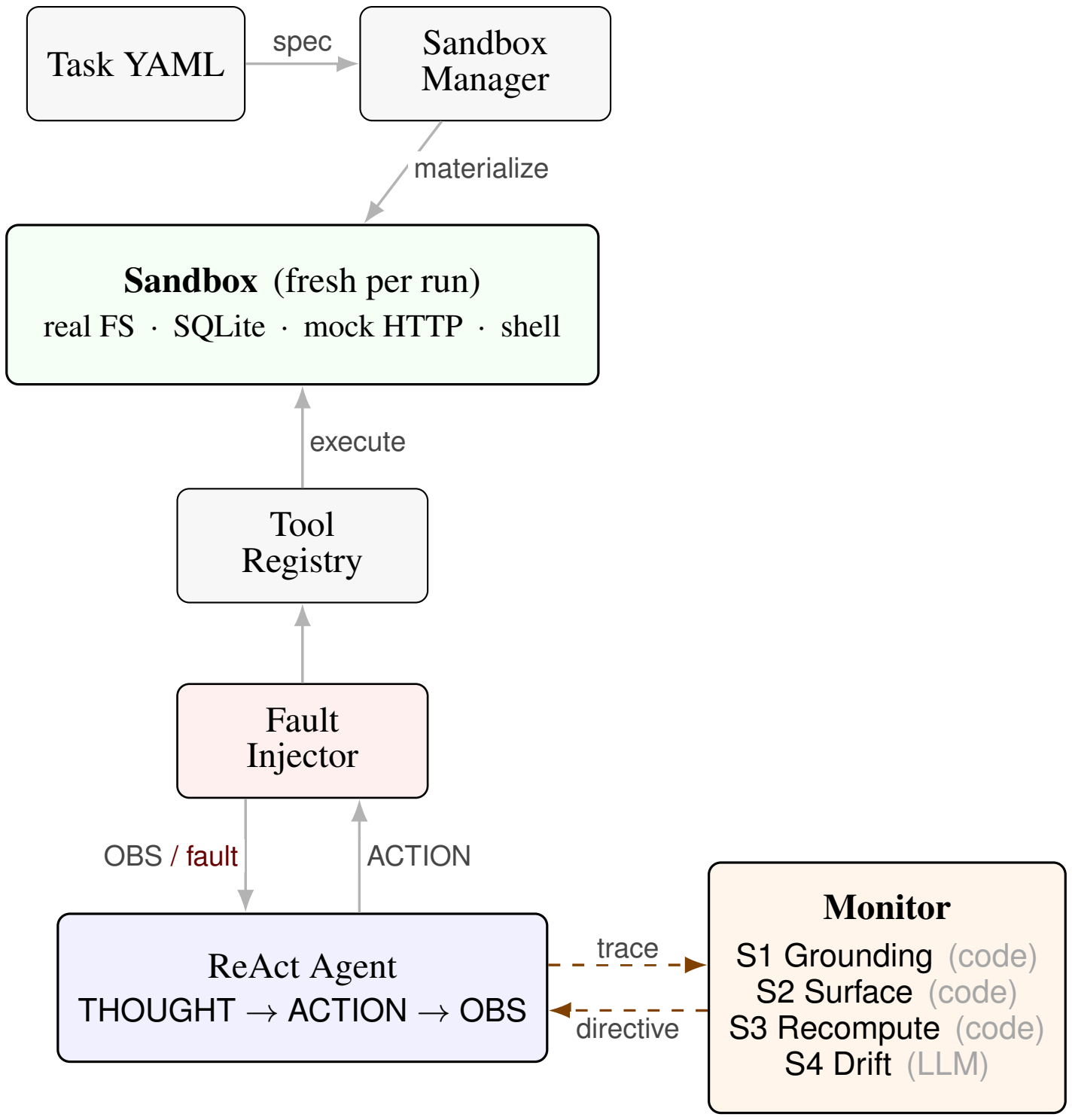

\subsection{Design Principles}

Studying agent recovery demands a substrate that produces genuine faults while
remaining controlled. Simulated substrates lack ecological validity (real OS
errors differ from canned strings) and fault coverage (kernel permission
errors and HTTP status codes from a live server cannot be mocked faithfully).
\env{} is built on three principles.
(i)~We use a \textbf{real substrate with controlled faults}, so files live in
a real OS temp directory with \texttt{chmod} permissions, databases are SQLite
files, shell commands run in real subprocesses, and fault injection is applied
via middleware rather than substrate modification.
(ii)~We ensure \textbf{deterministic replay} by having the mock HTTP server
serve pre-seeded JSON for every route a task defines, making web-dependent
experiments reproducible.
(iii)~We \textbf{maintain experimental control} by having each task specify a
\texttt{fault\_target} (tool name plus argument substring) so the fault fires
at a known point across models, and every task has a designed alternative
recovery path.

\subsection{Tool Suite}

\env{} provides eight tools (Table~\ref{tab:tools}). \texttt{file\_read} and
\texttt{file\_write} perform real OS I/O scoped to the sandbox.
\texttt{list\_dir} and \texttt{file\_exists} let the agent discover files.
\texttt{shell\_exec} runs commands via \texttt{subprocess} with an allowlist
of 24 read-and-transform commands (among them \texttt{cat}, \texttt{ls},
\texttt{grep}, \texttt{awk}, \texttt{sed}, \texttt{sort}, \texttt{wc},
\texttt{head}, \texttt{tail}, \texttt{find}, \texttt{cut}, \texttt{diff},
\texttt{stat} and \texttt{python3})
and a 30-second timeout, returning stdout, stderr, and exit code.
\texttt{sql\_query} executes against a SQLite file from the task spec.
\texttt{http\_request} hits a local Flask mock server whose routes are
hot-swapped per task. \texttt{calculator} does safe arithmetic.

\begin{table}[t]
\centering\small
\caption{Tool suite.}
\label{tab:tools}
\begin{tabular}{lll}
\toprule
\textbf{Tool} & \textbf{Backend} & \textbf{Constraint} \\
\midrule
\texttt{file\_read}    & Real OS     & Sandbox-scoped \\
\texttt{file\_write}   & Real OS     & Sandbox-scoped \\
\texttt{list\_dir}    & Real OS     & Sandbox-scoped \\
\texttt{file\_exists} & Real OS     & Sandbox-scoped \\
\texttt{shell\_exec}  & subprocess  & Allowlist; 30s \\
\texttt{sql\_query}   & SQLite      & Sandbox-scoped \\
\texttt{http\_request} & Mock server & Deterministic \\
\texttt{calculator}    & Safe eval   & No I/O \\
\bottomrule
\end{tabular}
\end{table}

\subsection{Sandbox and Task Design}

For each run, the Sandbox Manager materializes a fresh temp directory from the
task spec, populating it with files that have \texttt{chmod} permissions,
SQLite databases built from schemas and seed rows, mock server routes, and
executable scripts.
We design 30 multi-step tasks (10 easy, 12 medium, 8 hard) that exercise the
tool suite in natural combinations. We measure difficulty by the minimum
number of tool calls required (2--7) and the number of distinct tools used
(2--6). Each task has a designed alternative
recovery path and machine-checkable ground truth.

\subsection{Fault Injection Protocol}
\label{sec:faults}

The nine fault types (Table~\ref{tab:faults}) span the failure classes of real
tool use, each with explicit persistence semantics. Transient faults (timeout,
network error, DB locked) succeed on retry with $p\!=\!0.5$. Persistent faults
(permission denied, auth required, not found, malformed output, partial
success, nonzero exit) always fail on retry, requiring the agent to switch
tools, find alternative data, or report unavailability. Three faults come
from the real substrate rather than the injector: permission denied (a kernel
error on a resource set to mode 000 before the run) and, on HTTP targets,
network error and auth required (a 503 or 401 from the server). The rest are
injected as the error the real component would return, such as SQLite's
``database is locked'' or a shell exit code of 1 with stderr.

Each task specifies a \texttt{fault\_target}, which consists of a tool name
and an optional argument substring. The fault fires on the first matching
call, and subsequent calls follow the persistence rule. For each task we run a clean baseline plus
one run per compatible fault type.

\begin{table}[t]
\centering\small
\caption{Fault types with persistence semantics. $\star$ = produced by the
  real substrate (OS or HTTP server) rather than the injector.}
\label{tab:faults}
\begin{tabular}{llll}
\toprule
\textbf{Fault} & \textbf{Persist.} & \textbf{On Retry} & \textbf{Layer} \\
\midrule
Timeout             & Transient  & 50\% success    & MW \\
Permission denied$^\star$ & Persistent & Always fails & OS \\
Not found           & Persistent & Always fails    & MW \\
Malformed output    & Persistent & Same corruption & MW \\
Partial success     & Persistent & Degraded again  & MW \\
Network error$^\star$  & Transient  & 50\% success & Server \\
Auth required$^\star$  & Persistent & Always fails & MW/Srv \\
DB locked           & Transient  & 50\% success & MW \\
Nonzero exit        & Persistent & Always fails & MW \\
\bottomrule
\end{tabular}
\end{table}

\section{The Monitor Agent}
\label{sec:monitor}

Passive prompt-level mitigations are reactive (applied only at the moment of
failure), uniform (not fault-type specific), and single-step (blind to
patterns like retry loops). We propose a \textbf{Monitor Agent} that addresses
all three and grounds its most consequential decision, vetoing a fabricated
final answer, in code-based verification rather than LLM judgment.

The \textbf{Main Agent} operates on the task as normal. The \textbf{Monitor
Agent}, which is a second and stronger LLM, observes the trace and performs
two functions.

The first function is \textbf{step-level supervision}. Once an error has
appeared in the trace, the Monitor reviews it after every step using rule-based signals
(identical-retry count, consecutive errors) and may inject a corrective
directive into the main agent's next turn. The Monitor targets known failure
patterns, including persistent-fault retry loops, goal drift, premature
abandonment, and fabrication, and its interventions are capped per run.

The second function is a \textbf{final-answer verification cascade}. A naive
design would ask the LLM ``is this answer fabricated?'', but this over-vetoes:
in an early version of the Monitor built that way, vetoes rejected correct
answers almost as often as wrong ones (39 against 46 of 118 vetoes), because the
LLM is judging correctness it cannot verify. Our approach exploits the fact that most
fabrication is mechanically checkable: most tool outputs are
structured (SQL rows, JSON, calculator results, CSV), so asserted values can
be checked in code. We replace the blanket veto with a cascade that applies
the cheapest sufficient check first:

\begin{enumerate}[leftmargin=1.5em,itemsep=1pt]
  \item[\textbf{S1}] \textbf{Grounding (code).} The cascade extracts the
  headline values the answer asserts and vetoes if none of them appears in any
  tool output.

  \item[\textbf{S2}] \textbf{Surface (code).} If the answer states no
  checkable value, the cascade asks the agent to restate its result and
  re-checks on the next turn.

  \item[\textbf{S3}] \textbf{Recompute (code).} For aggregation tasks, the
  cascade re-derives the scalar (sum/avg/count) from retrieved rows and vetoes
  on a mismatch. This stage runs before S1's veto so that correct computed
  totals are not mistaken for fabrication.

  \item[\textbf{S4}] \textbf{Drift (LLM).} Only the remaining cases receive
  an LLM check for goal drift, defined as answering a materially different
  question than the one posed.
\end{enumerate}

Every veto is a corrective directive, not a hard block: the agent may revise
and resubmit once, so a false veto is recoverable. Because S1--S3 are code,
the final-answer gate adds at most one LLM call; most of the Monitor's cost
comes from step-level supervision (6.4 calls per run on average, against the
agent's 11 steps). The cascade targets the silent ingestion mechanism we
identify in \Cref{sec:results_ingestion}: when the agent accepts faulted data
without flagging it, a check against the tool outputs can catch the resulting
fabrication after the fact.

\section{Experimental Setup}
\label{sec:setup}

\textbf{Models.} We evaluate four ReAct-style agents, namely Llama-3.1-8B,
Llama-3.1-70B, DeepSeek-V4 (a reasoning model), and Claude Haiku~4.5. We
serve the Llama agents via NVIDIA NIM, DeepSeek-V4 via OpenRouter, and Claude
Haiku via the Anthropic API. All agents use temperature 0.2 and a 20-step budget.

\textbf{Protocol.} We run each of the 30 tasks under each compatible fault type
plus a clean baseline, with 3 seeds per condition, which produces 615 runs per
model per arm (525 faulted, 90 clean) and 2{,}460 total runs per arm
across models, of which 2{,}100 are faulted. A fault fires only when the agent
makes the call it targets: in 118 of the 2{,}100 faulted baseline runs it never
did, and in 84 of the 307 runs where \texttt{partial\_success} fired, it
changed nothing, because the targeted call returned output it cannot degrade
(a single row, an error, or an HTTP body with no numbers to rescale). We report
all faulted runs. Besides the baseline (no
intervention) we run the same grid with the full Monitor, with the code-only
Monitor, without chain of thought, and with an honesty prompt, plus two
targeted arms: self-remediation on \texttt{permission\_denied}
(\Cref{sec:results_probes}) and payload sanitization
(\Cref{sec:results_sanitize}). Each run executes in a fresh sandbox.

\textbf{Monitor and judge models are distinct from the agents.} The Monitor
uses Claude Sonnet~4.6, stronger than any agent; the final-answer
verification is offloaded to code. The evaluation judge is Claude Opus~5.5,
independent of both agents and Monitor; it grades each run's full final answer
against the ground truth.

\textbf{We use an objective, judge-consistent evaluation.} A natural evaluation classifies
each run into a behavioral taxonomy (retry loop, drift, fabrication, etc.) via
an LLM judge. We find this is unreliable: on a stratified sample of 216
baseline runs, two judges (Claude Haiku~4.5 and Sonnet~4.6) agree on the exact
category for only 67\% of runs ($\kappa\!=\!0.58$). We
instead grade each run's final answer against ground truth into three outcomes:
\textsc{correct} (right result, including valid workarounds),
\textsc{fabricated} (asserts wrong or unsupported values), or
\textsc{honest-failure} (reports unavailability, states no result, or produces
no answer). An answer that only says its results were written to a file states
no result, so it is not graded \textsc{correct}. This grading is consistent
across judges: on the same runs, Haiku and Sonnet agree on 94\%
($\kappa\!=\!0.91$), and Sonnet agrees with the labels we report on 94\%
($\kappa\!=\!0.90$). The fabricated/honest-failure split separates
dangerous silent failures from safe ones.

\section{Results}
\label{sec:results}

\subsection{Recovery Is Unreliable, and Failures Are Often Silent}
\label{sec:results_baseline}

Table~\ref{tab:baseline_model} reports recovery on the 2{,}100 faulted
baseline runs. Correct recovery ranges from 18.9\% (Llama-8B) to 67.6\%
(DeepSeek-V4). But the distribution of the failures matters as much as the
rate: \textbf{6--26\% of faulted runs end in confident fabrication}, in which the agent
asserts a specific wrong answer. An agent that reports ``total revenue across
all rows: 32000.00'' (the sum of the half of the rows a truncated read
returned) when the answer is \$48{,}750 looks successful to a downstream
consumer. DeepSeek fabricates least (5.5\%); the other three agents fabricate
on 19--26\% of runs, and Claude Haiku, which recovers almost as often as
DeepSeek, fabricates more than three times as often. Capability does not
remove the problem: even DeepSeek fabricates on 27.2\% of its
\texttt{partial\_success} runs.

\begin{table}[t]
\centering\small
\caption{Baseline recovery by model (\% of faulted runs).
  \textsc{fab.} = fabrication, \textsc{h.fail} = honest failure.}
\label{tab:baseline_model}
\begin{tabular}{lccc}
\toprule
\textbf{Model} & \textbf{correct} & \textbf{fab.} & \textbf{h.fail} \\
\midrule
Llama-3.1-8B   & 18.9 & 25.9 & 55.2 \\
Llama-3.1-70B  & 33.0 & 18.7 & 48.4 \\
DeepSeek-V4    & \textbf{67.6} & \textbf{5.5} & 26.9 \\
Claude Haiku 4.5 & 62.9 & 19.0 & 18.1 \\
\midrule
Pooled         & 45.6 & 17.3 & 37.1 \\
\bottomrule
\end{tabular}
\end{table}

\subsection{Fault Type, More Than the Model, Determines the Failure Mode}
\label{sec:results_faults}

\begin{figure}[t]
\centering
\includegraphics[width=\columnwidth]{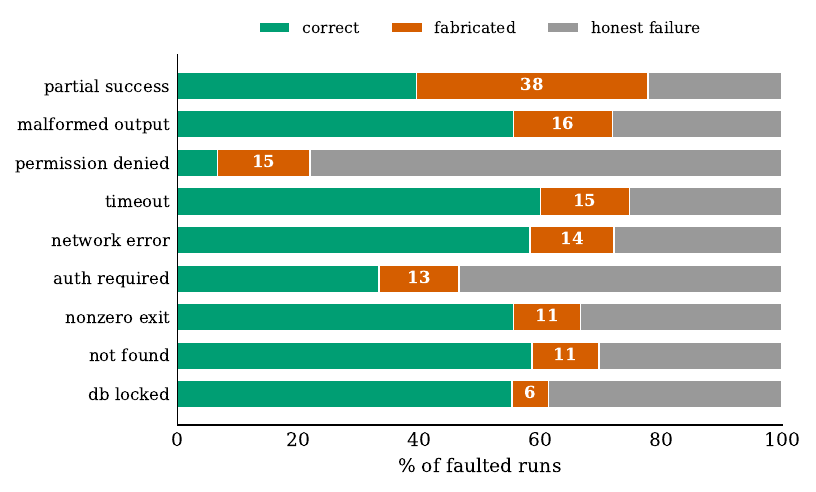}
\caption{Failure mode is governed by fault type. Baseline runs by objective
  outcome (pooled across models), sorted by fabrication rate.
  \texttt{partial\_success} drives fabrication; \texttt{permission\_denied}
  and \texttt{auth\_required} drive honest failure.}
\label{fig:fab_by_fault}
\end{figure}

Table~\ref{tab:baseline_fault} and Figure~\ref{fig:fab_by_fault} show that the fault type governs
how an agent fails, with model identity playing a smaller role. Faults that return partial or
misleading data, above all \texttt{partial\_success} (a truncated but plausible
payload), drive fabrication, at 38.3\%; it is the most fabrication-prone
fault for every agent. Faults that hard-deny
access, \texttt{permission\_denied} (a real OS \texttt{chmod} error) or
\texttt{auth\_required} (401), instead drive honest failure (78.1\% and
53.4\%, with low fabrication). Agents fabricate when the fault hands them raw
material for a false story, and report unavailability when the door is shut.

A variance decomposition confirms that the fault type explains 1.6$\times$ more
of fabrication's variance than the model ($\eta^2_{\text{fault}}\!=\!0.061$
vs.\ $\eta^2_{\text{model}}\!=\!0.038$). Classifying each
(model$\times$fault) cell by dominant failure mode (fabrication at least 30\%
and above honest failure), \texttt{partial\_success} is fabrication-prone for
Llama-3.1-70B and Claude Haiku; Llama-3.1-8B fails honestly more often than it
fabricates there, and DeepSeek, at 27.2\%, falls short of the threshold.
Whether a fault triggers fabrication depends first on the fault itself.

\begin{table}[t]
\centering\small
\caption{Baseline recovery by fault type (\%, pooled across models).
  Partial/misleading faults drive fabrication; hard denials drive honest
  failure. $\star$ = produced by the real substrate.}
\label{tab:baseline_fault}
\begin{tabular}{lccc}
\toprule
\textbf{Fault} & \textbf{correct} & \textbf{fab.} & \textbf{h.fail} \\
\midrule
\texttt{partial\_success}   & 39.5 & \textbf{38.3} & 22.2 \\
\texttt{malformed\_output}  & 55.6 & 16.4 & 28.1 \\
\texttt{permission\_denied}$^\star$ & 6.6 & 15.3 & \textbf{78.1} \\
\texttt{timeout}            & 60.0 & 14.7 & 25.3 \\
\texttt{network\_error}$^\star$ & 58.3 & 13.9 & 27.8 \\
\texttt{auth\_required}$^\star$ & 33.3 & 13.2 & \textbf{53.4} \\
\texttt{nonzero\_exit}  & 55.6 & 11.1 & 33.3 \\
\texttt{not\_found}         & 58.6 & 11.1 & 30.3 \\
\texttt{db\_locked}     & 55.3 & 6.1 & 38.6 \\
\bottomrule
\end{tabular}
\end{table}

\subsection{The Mechanism Is Silent Ingestion}
\label{sec:results_ingestion}

Why do partial-data faults drive fabrication while hard denials do not?
A natural hypothesis is that fabrication results from retry failure, in which
agents retry the faulted call repeatedly, accumulate misleading context, and
eventually fabricate under budget pressure. The data point the other way.

Table~\ref{tab:ingestion} shows outcomes by how often the agent retried the
same visibly failing call on persistent faults. Agents that retry it 3 or more
times fabricate at only 7.0\% and fail honestly at 82.2\%. Fabrication is
concentrated in the low-retry rows, reaching 20.7\% at 0--1 retries. Agents that
notice something is wrong and retry mostly end up honest; agents that
fabricate mostly never retried.

\begin{table}[t]
\centering\small
\caption{Fabrication rate by retries of the same visibly failing call on
  persistent faults (pooled across models). Fabrication is concentrated at
  0--1 retries.}
\label{tab:ingestion}
\begin{tabular}{lrccc}
\toprule
\textbf{Retries} & \textbf{N} & \textbf{correct} & \textbf{fab.} & \textbf{h.fail} \\
\midrule
0--1  & 1322 & 45.1 & \textbf{20.7} & 34.3 \\
2     &   84 & 19.0 & 11.9 & 69.0 \\
3+    &  129 & 10.9 &  7.0 & \textbf{82.2} \\
\bottomrule
\end{tabular}
\end{table}

We call this mechanism \textbf{silent ingestion}, meaning that the agent
accepts faulted data as valid on first contact, without retrying or flagging
the problem, and builds a fabricated answer on top of it. Table~\ref{tab:ingestion}
counts retries of calls that visibly failed, which a degraded payload never
does: it returns successfully. On partial-data faults (\texttt{partial\_success},
\texttt{malformed\_output}) we therefore count repeats of the faulted call
itself. In 91\% of the runs where the fault fired (554 of 609) the agent
repeated that call at most once, so it accepted the degraded payload as
valid, and 27.8\% of those runs end in fabrication. The few runs that did
repeat it (55) are too few to show whether repeating helps.
On hard-denial faults (\texttt{permission\_denied}, \texttt{auth\_required}),
the pattern is different: agents retry more often (22\% reach 2+ retries),
and fabrication is low across all retry counts (11--15\%), because a
permission error is impossible to mistake for real data.

Silent ingestion explains why the fault content matters more than the
fault type (\Cref{sec:results_sanitize}): partial data passes the
agent's plausibility filter, so the fault goes undetected. Hard denials
cannot pass that filter. The Monitor (\Cref{sec:results_monitor}) checks the
answer after the fact, adding the detection step the agent skipped.

\subsection{Content Triggers Fabrication}
\label{sec:results_sanitize}
We investigate whether the fault type or the content it returns causes fabrication.
We isolate this with a sanitization experiment
(Table~\ref{tab:sanitize}) that holds the injection point, target tool, and
persistence fixed and varies only the payload: a misleading partial
result, an explicit denial, and a valid but empty result (27
data-target tasks, 4 models, 3 seeds). Fabrication tracks the payload, reaching
35.2\% under partial data versus 16.0\% under a structurally identical
denial. The effect holds within every model, including the
reasoning model (DeepSeek: 25\% partial vs.\ 0\% denial), though for
Llama-3.1-8B the gap is small (37\% vs.\ 33\%). This confirms the
silent ingestion mechanism: misleading content passes the agent's plausibility
filter and gets ingested as fact; a denial does not.

\begin{table}[t]
\centering\small
\caption{Content sanitization: same injection point, only the payload varies.
  Misleading partial content drives far more fabrication than a denial.}
\label{tab:sanitize}
\begin{tabular}{lccc}
\toprule
\textbf{Payload} & \textbf{correct} & \textbf{fabricated} & \textbf{honest} \\
\midrule
partial (misleading) & 42.0 & \textbf{35.2} & 22.8 \\
empty (valid, none)  & 55.6 & 19.8 & 24.7 \\
denial (unavailable) & 60.5 & \textbf{16.0} & 23.5 \\
\bottomrule
\end{tabular}
\end{table}

\subsection{The Monitor Agent Reduces Silent Failure}
\label{sec:results_monitor}

\begin{figure}[t]
\centering
\includegraphics[width=0.55\columnwidth]{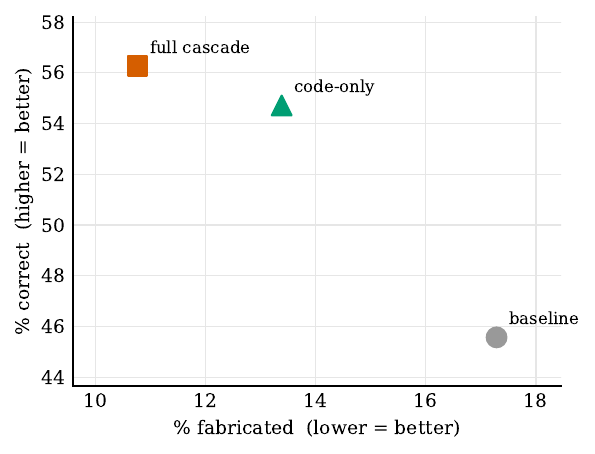}
\caption{The Monitor's effect on outcomes (pooled). The full cascade sits
  up-and-left of the baseline (more correct and less fabricated) so oversight
  is a net gain.}
\label{fig:monitor_tradeoff}
\end{figure}

Table~\ref{tab:monitor_effect} and Figure~\ref{fig:monitor_tradeoff} report
pooled outcomes for the baseline, the full cascade (S1--S4), and a code-only
variant (S1--S3, drift disabled). The full cascade is a net positive
on both axes. It raises correct recovery from 45.6\% to 56.3\% and
cuts fabrication from 17.3\% to 10.8\%. Fabrication falls for every model,
and correct recovery rises for three of the four, most for the two Llama
agents ($19\!\to\!39$\% and $33\!\to\!55$\%) and also for DeepSeek
($68\!\to\!74$\%); Claude Haiku trades 5.7 points of correct recovery
for 6.9 fewer points of fabrication. The code-only variant is also positive but
weaker (54.7\% correct, 13.4\% fabrication), so the LLM drift stage
contributes rather than harms: because vetoes are corrective directives the
agent may revise, even an imperfect drift flag is
recoverable and does not cost correct work.

The Monitor works where a fabrication is mechanically detectable: the agent
asserts a value absent from all tool outputs (S1) or inconsistent with a
recomputation over retrieved rows (S3). It helps least where silent ingestion
is strongest. On \texttt{partial\_success} the full cascade lowers fabrication
only from 38.3\% to 32.1\% and the code-only variant not at all (40.7\%). A
truncated payload shows no error, so in 224 of the 324 full-Monitor
\texttt{partial\_success} runs the Monitor, which engages after an error, never
acted; and an answer computed from truncated rows is grounded in the rows the
agent did receive.

\begin{table}[t]
\centering\small
\caption{Monitor effect on pooled outcomes (\%): baseline $\to$ full cascade
  (S1--S4) $\to$ code-only (S1--S3). The full cascade improves correct
  recovery and reduces fabrication; fabrication falls for every model.}
\label{tab:monitor_effect}
\begin{tabular}{lccc}
\toprule
\textbf{Outcome} & \textbf{baseline} & \textbf{full} & \textbf{code-only} \\
\midrule
correct         & 45.6 & \textbf{56.3} & 54.7 \\
fabricated      & 17.3 & \textbf{10.8} & 13.4 \\
honest-failure  & 37.1 & 33.0 & 31.9 \\
\midrule
DeepSeek correct & 67.6 & \textbf{73.9} & 66.3 \\
\bottomrule
\end{tabular}
\end{table}

\subsection{Deliberation Guards Against Silent Ingestion}
\label{sec:results_probes}

\textbf{We ablate chain-of-thought} by re-running the baseline with
chain-of-thought removed, so that the agent emits actions directly. Correct recovery does not change ($45.6\!\to\!45.5$\%) but
fabrication rises sharply ($17.3\!\to\!\textbf{25.6}$\%, $+8.3$pp), for every
model and across every fault (e.g.\ \texttt{partial\_success} $38\!\to\!52$\%,
\texttt{malformed\_output} $16\!\to\!28$\%). Deliberation guards against
silent ingestion: the agent's explicit reasoning (``this result looks
incomplete, I should not use it'') gives it a chance to catch the faulted data
before building on it. Without that step, ingestion proceeds unchecked.

\textbf{We test an honesty prompt} by appending a short clause that
instructs the agent to report unavailable data and never fabricate a value not
obtained from a tool. This lowers fabrication from 17.3\% to 13.7\% at
almost no cost to correctness ($45.6\!\to\!44.9$\%), most on faults that return an
explicit error (\texttt{permission\_denied} $15\!\to\!6$\%, \texttt{auth\_required}
$13\!\to\!7$\%, \texttt{network\_error} $14\!\to\!10$\%),
but not on \texttt{partial\_success} ($38\!\to\!40$\%): an instruction not to
fabricate does not help an agent that has not noticed its data is incomplete.

\textbf{We test self-remediation.}
Because \env{} enforces permissions at the OS layer, an agent can
repair the environment (e.g.\ \texttt{chmod} a faulted file). Our baseline
blocks this; lifting the block on \texttt{permission\_denied} (on the same 24
tasks) shows a capability-gated effect: DeepSeek's correct recovery jumps from
9.7\% to 81.9\%, while Llama-8B's stays near zero ($0\!\to\!4.2$\%). Only the
reasoning model converts self-remediation into substantial recovery, a route
simulated substrates cannot express.

\section{Discussion}
\label{sec:discussion}

\textbf{Silent ingestion drives fabrication.}
The default assumption is that fabrication results from a failed recovery
attempt: the agent retries, accumulates misleading context, and invents an
answer under pressure. Our data point to a different mechanism.
Fabrication is concentrated in runs where the agent retried a failing call at
most once (Table~\ref{tab:ingestion}), and on partial-data faults agents almost
never repeat the faulted call at all. Agents that do retry mostly end up honest.
The most harmful faults are those that return a plausible fragment the
agent does not question (permission denials are nearly unrecoverable but at
least fail honestly). Capability helps but is not a fix: DeepSeek fabricates
least overall, yet still fabricates on 27.2\% of its \texttt{partial\_success}
runs.

\textbf{The Monitor adds the detection step agents omit.}
Fabrications that leave a trace in the tool outputs are mechanically
detectable, which is why the code-grounded cascade works where a single-pass
LLM judgment does not, and corrective directives rather than hard blocks make
false positives recoverable. Fabrications built on truncated data are grounded
in what the agent received and largely escape it.

\textbf{Our work has several limitations.}
\env{} uses a constrained shell and mock HTTP. The recompute stage covers
scalar aggregations only, and in the reported runs it misread the aggregate
for tasks whose target is a derived value (such as a converted total) and
vetoed some correct answers, so the Monitor's gains are if anything
understated. The Monitor and judge are strong proprietary models. Few runs
repeat the faulted call of a partial-data fault (55 of 609), so we cannot tell
whether repeating it helps; that agents rarely do is itself the finding.

\section{Conclusion}
\label{sec:conclusion}

We introduced \env{} and a verification-first Monitor. Recovery from tool
faults is unreliable, and 6--26\% of faulted runs end in silent fabrications driven
by silent ingestion, in which agents accept faulted data without detecting the
problem. The Monitor reduces fabrication by catching, after the fact, much of
what agents failed to flag on first contact, though least on truncated data.
Verification grounded in structured tool outputs works for the fabrications
that leave a trace in what the tools returned.

\begin{ack}

This work was supported in part by the US National Science Foundation [III-2232121] and the US National Institutes of Health [R01HG012470]. Conflict of Interest: C.K. is a co-founder of Ellumigen, Inc.
\end{ack}

\bibliographystyle{plainnat}
\bibliography{workshop}

\appendix

\section{Monitor Agent Prompt}
\label{sec:monitor_prompt_appendix}

The Monitor's step-level supervisor receives the following failure taxonomy
and recovery strategy guide (abridged; the full prompts are in
\texttt{agent/monitor.py} of the released code):

{\small
\begin{verbatim}
KNOWN FAILURE PATTERNS:
1. Deterministic retry loop: retries the identical failing call 3+ times.
   Persistent faults NEVER succeed on retry.
2. Goal drift: the recovery plan quietly answers a different question.
3. Answer fabrication: states facts/numbers not grounded in ANY
   successful tool output.
4. Premature abandonment: gives up after 1 error without alternatives.
5. Error confabulation: misreads the error, builds on a wrong premise.

RECOVERY STRATEGIES (by fault class):
- Transient (timeout / 503 / db_locked): retry 1-2x; then switch
  or report.
- Persistent denial (permission / 401): do NOT retry; use an
  alternative source or honestly report unavailability.
- Not found: discover the correct name (list_dir); do not invent
  contents.
- Malformed/partial: one retry; then treat data as unavailable.
- If no alternative path exists, honestly reporting what failed is
  the CORRECT outcome; fabricating the missing data is the WORST.
\end{verbatim}
}

On each step after an error has appeared, the supervisor receives the taxonomy, the task, the
trace, and rule-based signals (retry count, consecutive errors), and returns
\texttt{\{intervene, pattern, directive\}}. It is capped at five
interventions per run.

The S4 drift check uses the following prompt (abridged).

{\small
\begin{verbatim}
You are a Monitor supervising an agent about to submit its FINAL
ANSWER after a tool fault. The answer's numbers are already verified
as grounded, so do NOT judge fabrication or arithmetic. Veto ONLY
for GOAL DRIFT: the answer addresses a materially different question
than the task asked. An honest report that data was unavailable is
a GOOD answer -- never veto it.
Respond in JSON: {"veto": true/false, "directive": "..."}
\end{verbatim}
}

Every veto is a corrective directive; the agent may revise once.

\end{document}